\documentclass[preprint,amsmath,amssymb,keywords,showkeys,superscriptaddress,nofootinbib]{revtex4}
\usepackage{graphicx}
\usepackage{lipsum}
\usepackage[utf8]{inputenc}
\usepackage[english]{babel}
\usepackage{graphicx}
\usepackage{bm}
\usepackage{adjustbox}
\usepackage{multirow}

\begin{document}

\begin{abstract}
Raman spectroscopy in combination with machine learning has significant promise for applications in clinical settings as a rapid, sensitive, and label-free identification method. These approaches perform well in classifying data that contains classes that occur during the training phase. However, in practice, there are always substances whose spectra have not yet been taken or are not yet known and when the input data are far from the training set and include new classes that were not seen at the training stage, a significant number of false positives are recorded which limits the clinical relevance of these algorithms. Here we show that these obstacles can be overcome by implementing recently introduced Entropic Open Set and Objectosphere loss functions. To demonstrate the efficiency of this approach, we compiled a database of Raman spectra of 40 chemical classes separating them into 20 biologically relevant classes comprised of amino acids, 10 irrelevant classes comprised of bio-related chemicals, and 10 classes that the Neural Network has not seen before, comprised of a variety of other chemicals. We show that this approach enables the network to effectively identify the unknown classes while preserving high accuracy on the known ones, dramatically reducing the number of false positives while preserving high accuracy on the known classes, which will allow this technique to bridge the gap between laboratory experiments and clinical applications. 

\end{abstract}

\title{Raman spectroscopy in open world learning settings using the Objectosphere approach}

\author{Yaroslav Balytskyi}
\email{ybalytsk@uccs.edu}
\affiliation{Department of Physics and Energy Science, University of Colorado Colorado Springs, Colorado Springs, CO, 80918, USA}
\affiliation{UCCS BioFrontiers Center, University of Colorado Colorado Springs, Colorado Springs, CO, 80918, USA}

\author{Justin Bendesky}
\affiliation{UCCS BioFrontiers Center, University of Colorado Colorado Springs, Colorado Springs, CO, 80918, USA}
\affiliation{Department of Chemistry, New York University, New York, NY 10003, USA}

\author{Tristan Paul}
\affiliation{Department of Physics and Energy Science, University of Colorado Colorado Springs, Colorado Springs, CO, 80918, USA}
\affiliation{UCCS BioFrontiers Center, University of Colorado Colorado Springs, Colorado Springs, CO, 80918, USA}

\author{Guy Hagen}
\affiliation{UCCS BioFrontiers Center, University of Colorado Colorado Springs, Colorado Springs, CO, 80918, USA}

\author{Kelly McNear}
\affiliation{Department of Physics and Energy Science, University of Colorado Colorado Springs, Colorado Springs, CO, 80918, USA}
\affiliation{UCCS BioFrontiers Center, University of Colorado Colorado Springs, Colorado Springs, CO, 80918, USA}

\date{\today}


\maketitle

\section*{Introduction}

Bacterial infections are responsible for severe diseases and cause 7 million deaths annually~\cite{fleischmann2016assessment,deantonio2016epidemiology}. The conventional methods for bacteria detection in clinical applications, enzyme-linked immunosorbent assays (ELISA), and strategies based on the polymerase chain reaction (PCR) and sequencing take significant time~\cite{howes2014plasmonic,chung2013magneto,levy2018surviving,chaudhuri2008efns}. Clinical diagnostic results are often delayed up to 48 hours for microbiological culture with an additional 24 hours for antibiotic susceptibility testing. While waiting for the results, broad-spectrum antibiotics (BSAbx) are typically applied until a specific pathogen is identified. Even though BSAbx are often lifesaving, they contribute to the rise of antibiotic resistant strains of bacteria and decimate the healthy gut microbiome, leading to overgrowth of pathogens such as \textit{Candida albicans} and \textit{Clostridium difficile}~\cite{brusselaers2011rising,sullivan2001effect}, and according to the Centers for Disease Control and Prevention, over $30\%$ of patients are treated unnecessarily~\cite{fleming2016prevalence}. As a result, delays in pathogen identification ultimately lead to longer hospital stays, higher medical costs, and increased antibiotic resistance and mortality~\cite{davies2010origins}. There is a clear need for new culture-free, rapid, cost-effective and highly sensitive detection and identification methods to ensure that targeted antibiotics are applied as a means to mitigate antimicrobial resistance. 

Raman spectroscopy is a non-destructive technique that has the ability to meet these requirements by providing a unique, fingerprint-like spectrum of a sample. It consists of inelastic light scattering from a sample and the successive measurement of the energy shift of scattered photons by the detector~\cite{pang2016review}. While Raman spectroscopy is a highly sensitive technique capable of identifying a wide variety of chemical species, analyzing the spectra can be difficult and data processing is necessary to identify species with the accuracy needed for clinical applications. Due to the large amounts of data which is acquired, it is difficult and impractical to hard-code all rules of classification. Automating this process requires some form of Machine Learning (ML) which represents a code being capable of improving itself by learning the relevant features from the data rather than hard-coding them. After the new knowledge is acquired, it can be used to tackle real-world problems such as performing classification or regression~\cite{goodfellow2016deep}. 

Machine learning has proven itself to be a powerful tool in multiple applications. For example, shallow ML models, such as logistic regression, can be employed in some situations, but in order for such shallow models to work, one has to choose the relevant features manually, i.e. perform feature engineering to enable the classification operation. In other words, the relevant features have to be fed into the model manually. To overcome this limitation, deep learning (DL) models using neural networks (NN) have been developed, allowing models to explore very abstract features with little human intervention. Deep learning models with various architectures were applied with great success in image processing ~\cite{krizhevsky2012imagenet,farabet2012learning,szegedy2015going,Hagen2021} as well as speech ~\cite{dahl2013improving, hinton2012deep} recognition. Not only this, but DL methods can be applied to a variety of fields; from medicine ~\cite{goodacre1998rapid,sigurdsson2004detection,olaetxea2020machine} and food safety~\cite{marigheto1998comparison} to particle physics~\cite{ciodaro2012online}. It was later found that DL methods can effectively distinguish the spectra of different molecules~\cite{liu1993chemometric} and do so more effectively than linear regression methods for data analysis in Raman spectroscopy~\cite{gniadecka1997diagnosis}. This demonstrates that these methods are ideal to explore for our goals. DL methods using convolutional neural networks were shown to boost the accuracy in comparison with other types of dense networks~\cite{liu2017deep}, which can suffer from a vanishing gradient problem, preventing increases in accuracy~\cite{hochreiter2001gradient}. The ResNet architecture overcomes this problem and boosts the accuracy even further by introducing skip connections between the layers~\cite{he2016deep}. Current state-of-the art deep learning algorithms for Raman spectroscopy favor the ResNet architecture because it can maintain high performance at a lower complexity than its competitors. Using these advances, Raman spectroscopy combined with machine learning has shown significant promise for clinical use as a rapid, label-free identification method~\cite{ho2019rapid,maruthamuthu2020raman,thrift2020deep,lussier2020deep,peiffer2020machine,lu2020combination}. 

 However, a significant limitation of the aforementioned NNs are that they are only tested on the classes on which they were initially trained. As a result, if the inputs for the NN are far from the training set and include new, never seen before classes, the NN's behavior becomes ill-defined. Simply put, if the NN is trained to identify either ``cats" or ``dogs" but during testing sees a new class of ``fish", the false-positive rate becomes uncontrolled and the NN will misclassify ``fish" as either ``cat" or ``dog". When we consider real-world settings and practical applications, collecting data of all possible chemical or pathogenic species is not feasible, so there will always be unknown classes. To avoid false-positives and misclassification, the ideal NN should be able to not only classify the known samples, but also be able to handle any unknowns without misclassification. This obstacle is a significant limitation in the currently deployed ML systems and represents a gap between the developments of novel identification methods and their adoption into routine clinical practices.

To overcome this obstacle, we implement ideas from Entropic Open-Set and Objectosphere approaches~\cite{dhamija2018reducing} and combine them with the ResNet26 architecture to show, for the first time, that ResNet architecture can be modified to accurately and reliably reject samples it has not seen before while remaining highly accurate in its identification of the known classes. This new combined architecture significantly outperforms the naive approaches such as the thresholding softmax score ~\cite{matan1990handwritten,de2000reject,fumera2002support} and background class methods~\cite{matan1990handwritten}. The general idea of implementing these approaches is to teach the NN to ignore the features of the irrelevant classes and only focus on the features of the relevant classes. To demonstrate the efficiency of our approach, we compiled a database of spectra of 40 chemical classes and separated them into distinct categories. Twenty of them are amino acids, the building blocks of proteins. The remaining 20 classes were split into two groups of 10 for the purpose of testing the classification accuracy of the NN. The spectra of the 20 amino acids were categorized as “known”, and the remaining 20 chemicals were randomly split into two groups we called ``ignored" or ``never seen before". The samples from the ``never seen before” class are unique in that the NN was not trained on them and does not see them until the testing stage. For our purposes, we introduced the following conventions of ``known classes”, ``ignored classes” and ``never seen classes”:
$$
\begin{cases}
\mathcal{K} \ - \ known \ classes \ or \ classes \ of \ interest  \\
\mathcal{I} \ - \ ignored \ classes   \\
\mathcal{N}  \ - \ never \ seen \ before \ classes  
\end{cases}
$$
During the training stage, the NN is taught to be highly accurate on the $\mathcal{K}$ classes and separate them from the $\mathcal{I}$ classes. During the test stage, we test its accuracy on $\mathcal{K}$ as well as determine the rate number of false positives of $\mathcal{N}$. 

\section*{Data collection and ResNet26 architecture}
\label{DataColl}

The 20 simple amino acids were purchased from Carolina Biological Supply Company (Burlington, NC). All other chemicals were purchased from either Alfa Aesar or Sigma Aldrich. All samples were measured in their crystalline form on a glass-bottomed petri dish. Raman maps were collected on a WITec alpha300 microscope in the confocal Raman mode, schematically shown in Fig.~\ref{Setup}. The excitation source was a 75 mW, 532 nm laser, focused with a Zeiss EC-Epiplan 20x 0.4 NA objective. A WITec UHTS 300 spectrometer with a 600 groove/mm grating was used to collect the spectra. The detector was an Andor DV401-BV spectroscopic CCD. This spectrometer has a resolution of 3 cm\textsuperscript{-1} or 0.009 nm. Raman maps were collected over an area of 50 $\mu$m by 50 $\mu$m. This area was divided into 100 x 100 pixels, each containing a Raman spectrum collected with an integration time of 0.1 seconds. Of the collected spectra, $\frac{5}{6}$ were used for training and $\frac{1}{6}$ were used for testing. Before feeding the data into the NN, the spectra are normalized, as shown in Fig.~\ref{Represent}. To do this, the large Rayleigh peak from the laser is cut and the subsequent spectrum is rescaled into a $\left[0, 1\right]$ range.

\begin{adjustbox}{center,caption={Schematic drawing of our experimental setup. The incident 532 nm green laser interacts with the sample on a glass slide and the scattered light (red) is collected by the dectector.},label={Setup},nofloat=figure,vspace=\bigskipamount}
\includegraphics[width=12 cm]{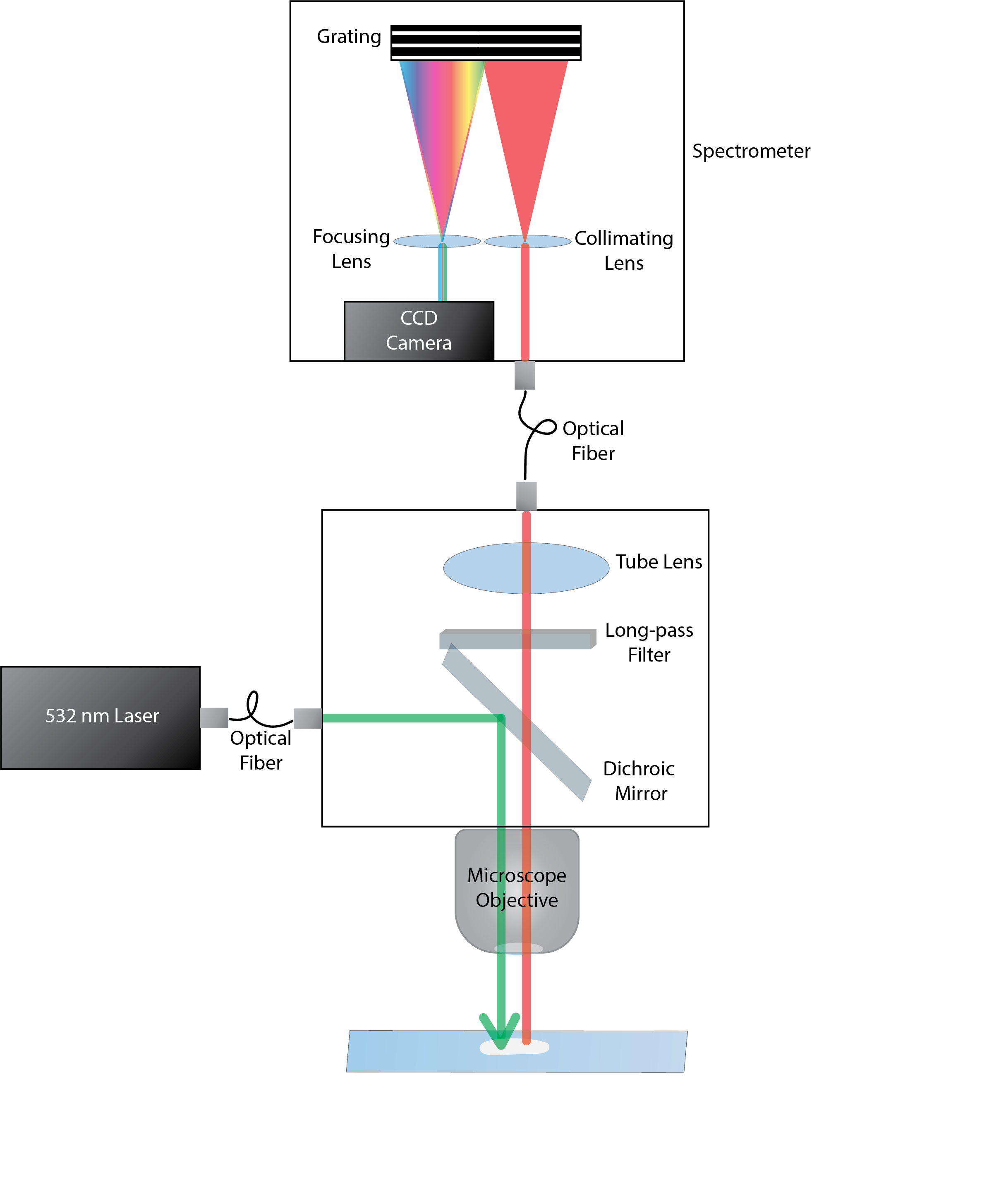}
\end{adjustbox}

\begin{adjustbox}{center,caption={Representative spectra for (a) raw and (b) normalized lysine data},label={Represent},nofloat=figure,vspace=\bigskipamount}
\includegraphics[width=\textwidth]{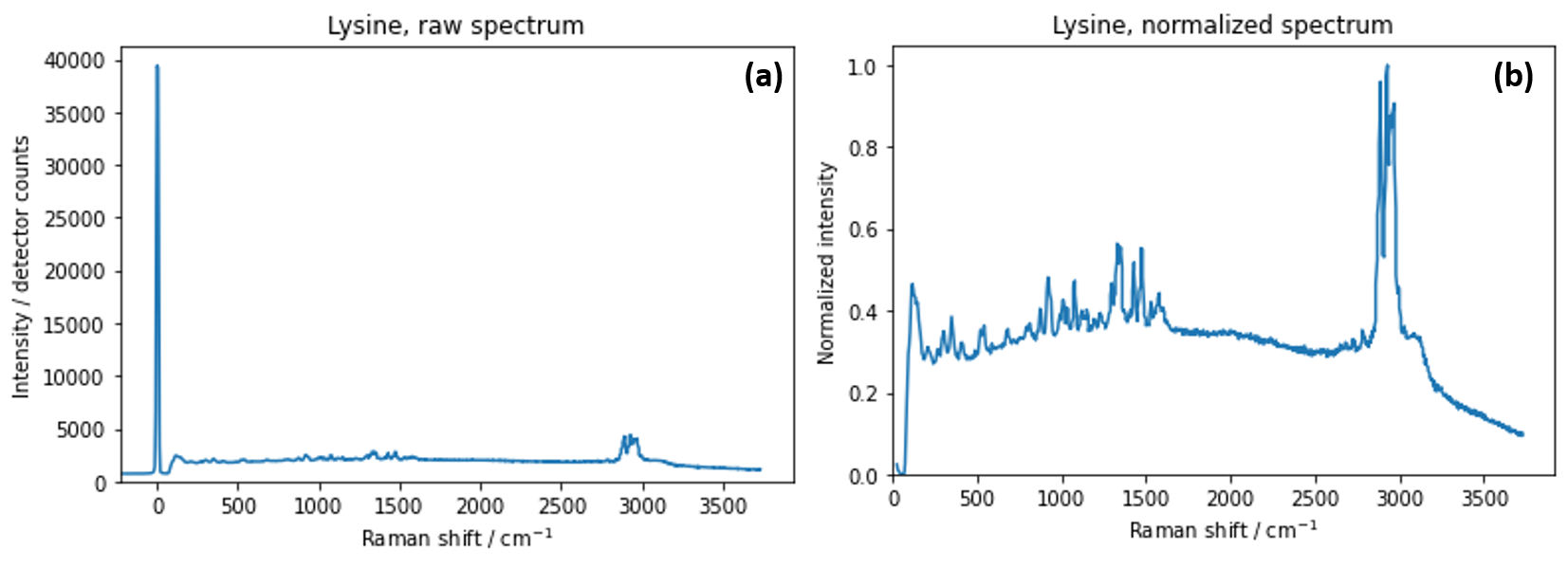}
\end{adjustbox}

Once the spectra were collected, they were categorized as either “known”, “ignored”, or “never seen” and are detailed in Table \ref{tab:table1}.

\begin{table}[h!]
    \begin{tabular}{|c|l|}
    \hline
      \textbf{    Class    } & \textbf{    Chemical Name  }\\
      \hline
      \multirow{5}{*}{$\mathcal{K}$} & DL-alanine, DL-aspartic acid, DL-isoleucine, DL-leucine,DL-methionine,\\
      
      & DL-phenylalanine, DL-serine, DL-threonine, DL-tyrosine, DL-valine,\\
      
      & glycine, L-arginine, L-asparagine, L-cystine, L-glutamic acid, L-histidine,\\
      
      &  L-lysine, L(+)-cysteine, L-proline, L-tryptophan
      \\
      \hline
      \multirow{2}{*}{$\mathcal{I}$} & anthrone, beta estradiol, chloroquine, fluconazole, laminarin, lauric acid,  \\
      
      & MOPS, methyl viologen, progesterone, uridine
     \\
      \hline
      \multirow{3}{*}{$\mathcal{N}$} & ampicillin, CHAPS, D(+) maltose monohydrate, forskolin,\\
      
      & 1-Decanesulfonic acid, polyvinyl pyrrolidone, potato starch,\\
      
      & sodium deoxycholate, sodium dodecylsulfate, silver nitrate
      \\
      \hline
    \end{tabular}
    \caption{Table of chemical names for known ($\mathcal{K}$), ignored ($\mathcal{I}$), and never seen before ($\mathcal{N}$) class identifications.}\label{tab:table1}
\end{table}

First, to make sure that our NN has sufficiently high accuracy in the ``closed-world" settings, we expose it to all classes, $\mathcal{K}$, $\mathcal{I}$ and $\mathcal{N}$. The conventional ResNet26 architecture shown in Fig.~\ref{ResNet26} is run 5 times as a means to help boost the accuracy of the network. While a few misclassifications may occur during each separate run, we can average the predictions (e.g. majority voting), and boost the accuracy to $99.99\%$ as shown in Fig.~\ref{Together40}. The accuracies for each run taken separately and the majority vote accuracy are the following:
$$
\begin{cases}
Accuracy\left[1\right] = 98.72 \% \\
Accuracy\left[2\right]  = 99.28 \% \\
Accuracy\left[3\right]  = 99.06 \% \\
Accuracy\left[4\right]  = 99.98 \% \\ 
Accuracy\left[5\right]  = 97.49 \%
\end{cases}\Rightarrow \underbrace{Final \ accuracy = 99.99\%}_{PredictionEnsemble = \frac{1}{5}\sum_{i = 1}^5 Prediction\left[i\right]}
$$
\begin{adjustbox}{center,caption={Schematic representation of the ResNet26 architecture. The processed spectra are fed into the NN in order to output classification probabilities.},label={ResNet26},nofloat=figure,vspace=\bigskipamount}
\includegraphics[width=13 cm]{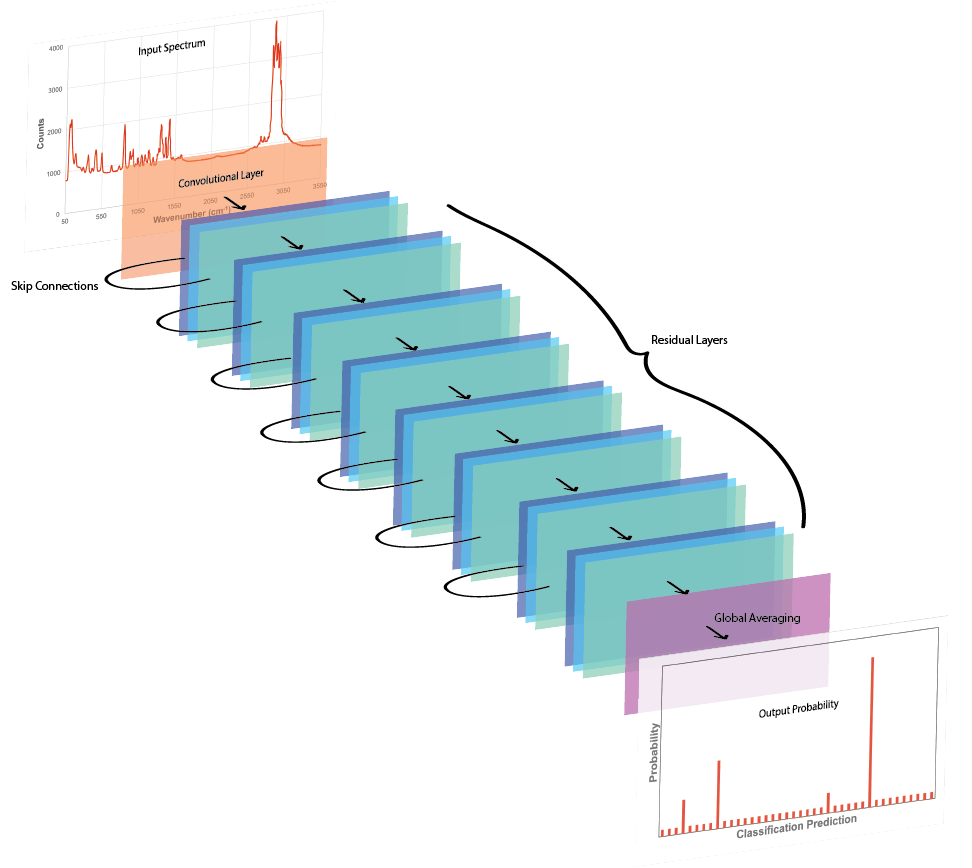}
\end{adjustbox}

\begin{adjustbox}{center,caption={Correlation table after the majority vote of 5 runs is taken. Accuracy = $99.99\%$.},label={Together40},nofloat=figure,vspace=\bigskipamount}
\includegraphics[width=\textwidth]{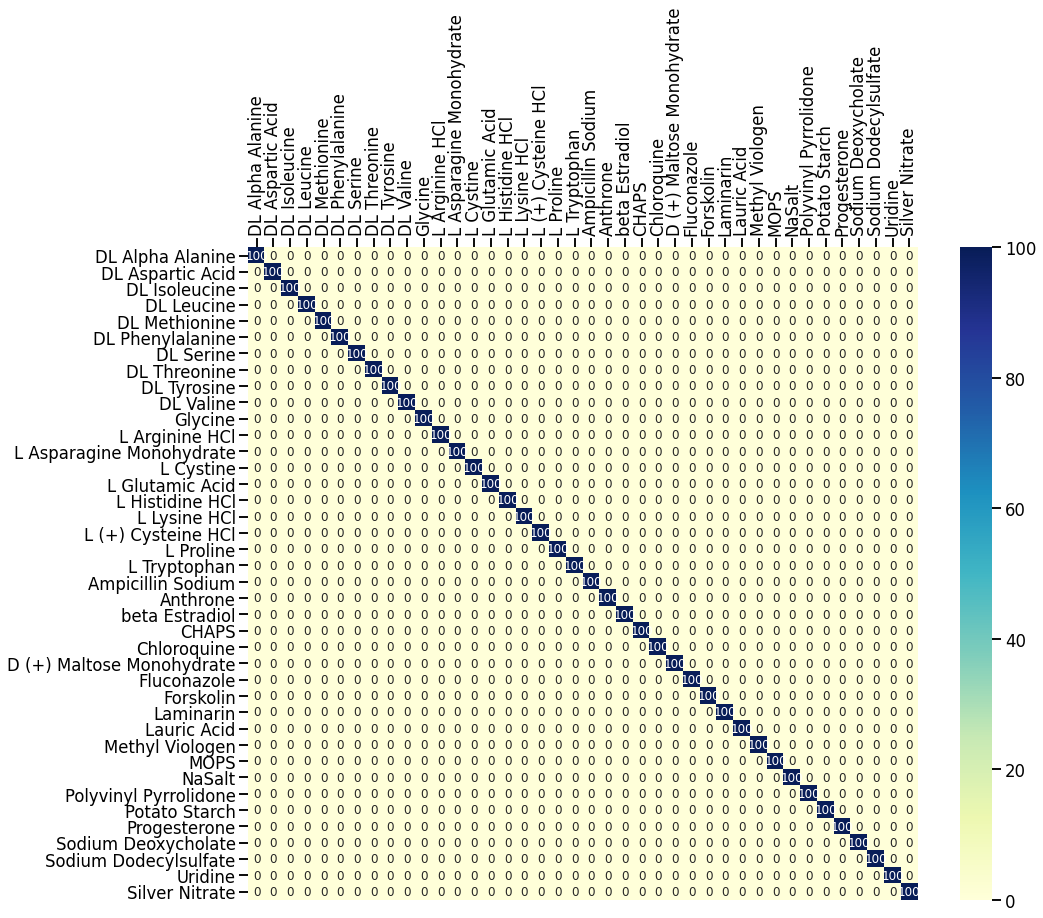}
\end{adjustbox}
However, as previously discussed, there will always be unknowns in real-word settings. Though the NN performs well and with high accuracy in ``closed-world" settings, it is not realistic. To demonstrate how this approach performs in the presence of unknowns, we perform the learning in the ``open world" settings. 

Here we concentrate on $\mathcal{K}$ and choose to ignore $\mathcal{I}$, effectively forcing the NN to expect the unexpected while keeping high accuracies on the known classes. The $\mathcal{N}$ classes are presented only during the testing stage. In the following sections, we investigate four different open-world approaches, \textbf{(1)} thresholding of the softmax score, \textbf{(2)} background class, \textbf{(3)} entropic open set, and  \textbf{(4)} Objectosphere, and demonstrate the advantages and limitations of each approach. We show that while all these approaches are highly accurate on the known classes, having $Accuracy = 99.97\pm 0.02 \%$, the Objectosphere approach is much more efficient in treating the $\mathcal{N}$ classes and is better prepared for unexpected inputs.

\section*{Background Class and Thresholding the Softmax Score}
\label{Naive}

First, we review the necessary background and terminology associated with neural networks. The NN has an initial input, $x$, which in our case corresponds to the values of the intensities in the spectra. The output corresponding to each class is $l_c\left(x\right)$. The deep feature of the  NN, $F$, is the output of the layer of the NN before the last one, and the output is obtained by multiplication by the weights $W$ to the last layer:
    $$l_c(x) = W \cdot F\left(x\right)$$
The final probability of which class, $c$, that a particular input will belong to is calculated by the ``softmaxing" procedure:
$$
S_c\left(x\right) = \frac{e^{l_c\left(x\right)}}{\sum_{c} e^{l_c\left(x\right)}}
$$
which lies in the range $S_c\left(x\right) \in \left[0; 1\right]$, and $\sum_c S_c\left(x\right) = 1$ and therefore is interpreted as a probability. The input sample will be classified as whichever class has the highest output probability.

The output of the NN is high-dimensional, so to provide a reasonable visualization, we use the Uniform Manifold Approximation and Projection (UMAP) dimensionality reduction technique~\cite{mcinnes2018umap}. This allows for the high-dimensional output of the NN to be visualized in a 2-dimensional way. In all our runs, we used the default parameters of the algorithm. In a 2D plane, different classes correspond to different colors, and should be separated from each other if the NN is highly accurate. In the case that there is any overlap of points of different colors, this indicates a mistake of the NN. As an example, in the ``closed-world" settings with 40 classes, the UMAP output is provided in Fig.~\ref{40UMAP}. Each class of the 40 classes correspond to a color in the gradient shown and we observe good separation between them and note that each color is concentrated in a particular point of a 2D space rather than being scattered over all the plane, a good indication that the NN is classifying efficiently. As we shift to the ``open-world" learning setting, the $\mathcal{K}$ classes correspond to the red to blue gradient. When new, never seen before classes are present during the testing stage, we group these unknowns together and represent them as violet dots on the UMAP diagrams.

\begin{adjustbox}{center,caption={UMAP of 40 classes in the closed world setting.},label={40UMAP},nofloat=figure,vspace=\bigskipamount}
\includegraphics[width=14 cm]{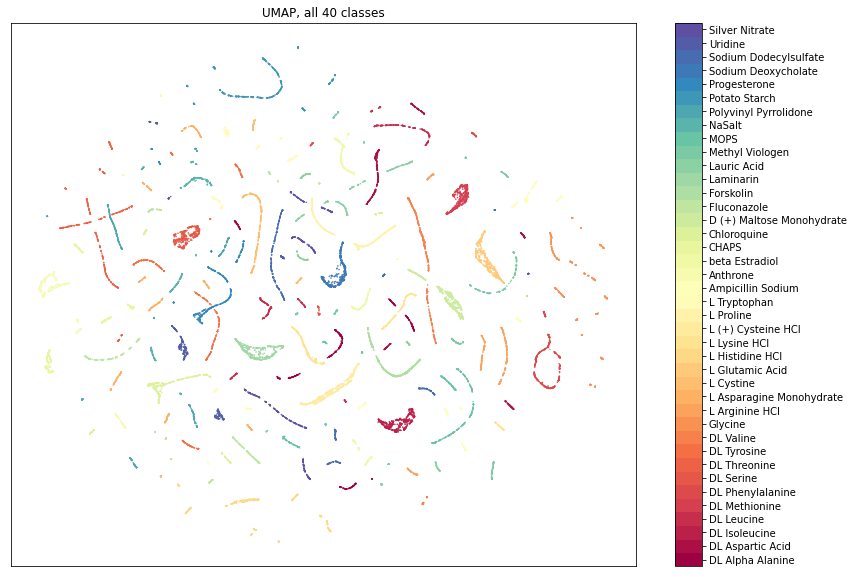}
\end{adjustbox}

When the unknown, never seen before classes are present, one naive approach is to include a trash or background class which corresponds to the background class method. The amino acid number $i\in\left[0,C - 1\right]$, and the $\{\mathcal{I, N}\}$ classes are encoded as:

$$\left(Amino \ acid\right)\left[i\right] = \underbrace{\left[0, \cdots,\underbrace{1}_{ith \ position}, \cdots, 0 \right]}_{Length = C + 1}; \ \mathcal{N} = \underbrace{\left[0, \cdots, \underbrace{1}_{20th \ position}\right]}_{Length = C + 1},$$ where $C = 20$ - number of classes of interest (amino acids).

Another naive approach is to threshold the softmax score, which assumes that the $\mathcal{K}$ and $\{\mathcal{I, N}\}$ classes are sufficiently separated in the feature space. In other words, Shannon's entropy of NN's output on the $\{\mathcal{I, N}\}$ classes is close to the maximal value $\log_2\left(C\right)$:
$$
\{\mathcal{I, N}\} = \underbrace{\left[\frac{1}{C}, \cdots, \frac{1}{C} \right]}_{Length = C}
$$
while the entropy of the output of $\mathcal{K}$ classes is close to zero. Assuming that this condition is fulfilled, one can introduce a cutoff, and if the maximum value of the softmax score is lower than its value, the input is classified as the new, never seen before class. 

However, as can be seen in Fig.~\ref{BackThSmUMAP}, both the background class and thresholding the softmax score methods---despite having a high accuracy on the known classes---work poorly if the $\mathcal{N}$ classes are present and significantly mixes the knowns and unknowns. As illustrated in Fig.~\ref{ThrThr}, even in a case of introducing a high cutoff value of $\Lambda = 0.99$ (corresponding to the NN being confident by $99\%$), the false positive (${FP}$) rate is  10.0\%. Additionally, on the $\mathcal{K}$ classes, in  $7.0\%$ of the cases the NN produces an inconclusive output, i.e. classifies $\mathcal{K}$ to be belonging to the $\mathcal{N}$ classes.

\begin{adjustbox}{center,caption={UMAP diagrams of (a) the background class and (b) the thresholding the softmax score methods. Class $\mathcal{N}$ (violet) is scattered over a large part of the diagram, mixing the known and the unknown.},label={BackThSmUMAP},nofloat=figure,vspace=\bigskipamount}
\includegraphics[width=15 cm]{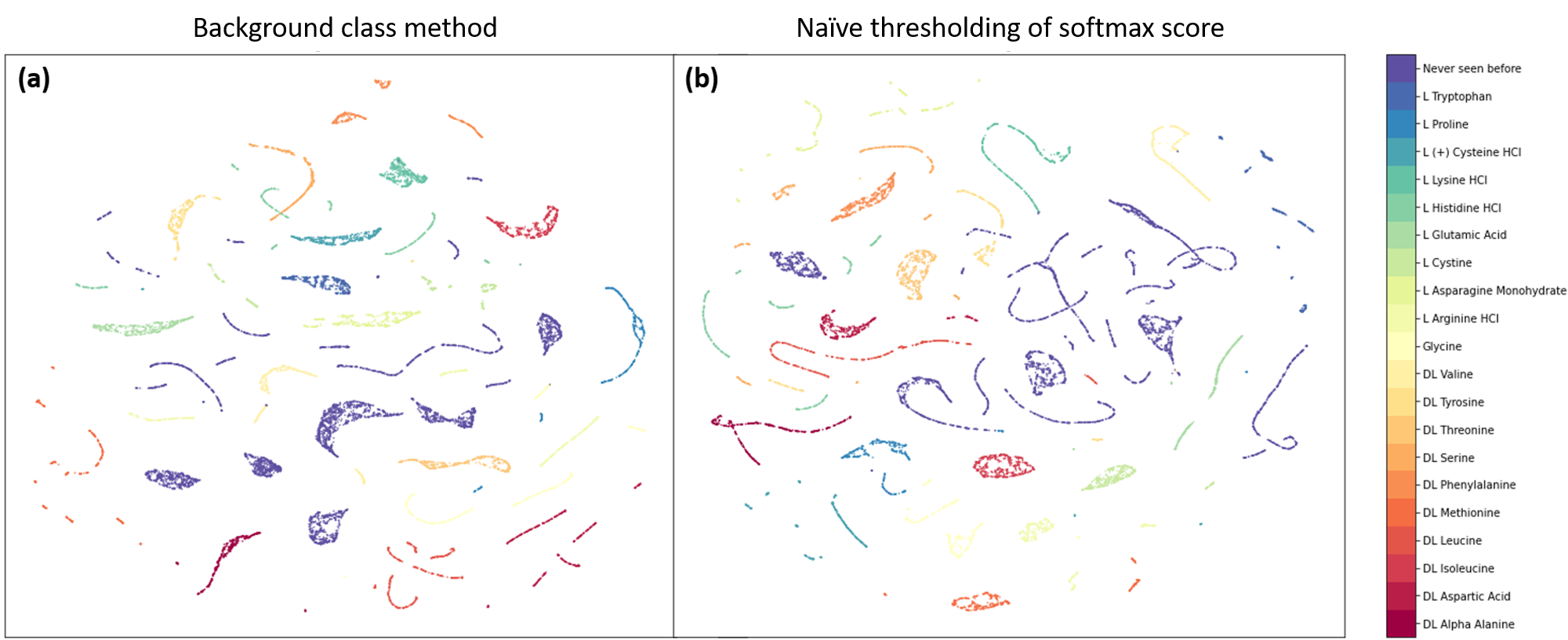}
\end{adjustbox}

\begin{adjustbox}{center,caption={Even at the large cutoff value $\Lambda = 0.99$, the $FP=10.0\%$, and the number of inconclusive outcomes on $\mathcal{K}$ classes is $7.0\%$.}, label={ThrThr},nofloat=figure,vspace=\bigskipamount}
\includegraphics[width=15 cm]{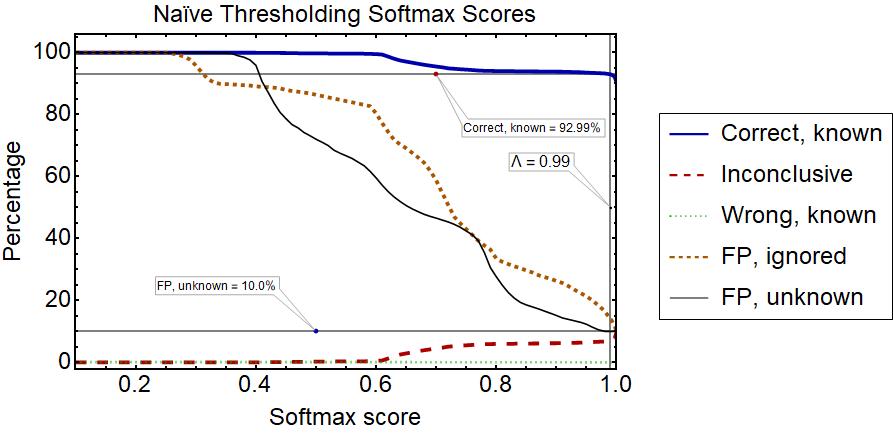}
\end{adjustbox}

Though the absolute values of the deep features $\lvert\lvert F \lvert\lvert$ of the known  tends to be slightly higher than those of ignored and unknowns as shown in Fig.~\ref{ThrFeat}, the amount of overlap between the three classes indicates that the NN has difficulty distinguishing between $\mathcal{K}$ and $\mathcal{N}$. The Entropic Open Set and Objectosphere approaches aim to increase this separation.

\begin{adjustbox}{center,caption={The absolute value of the deep feature $\lvert\lvert F \lvert\lvert$ of $\mathcal{K}$ classes tends to be higher than those of $\mathcal{N}$ and $\mathcal{I}$.}, label={ThrFeat},nofloat=figure,vspace=\bigskipamount}
\includegraphics[width=14 cm]{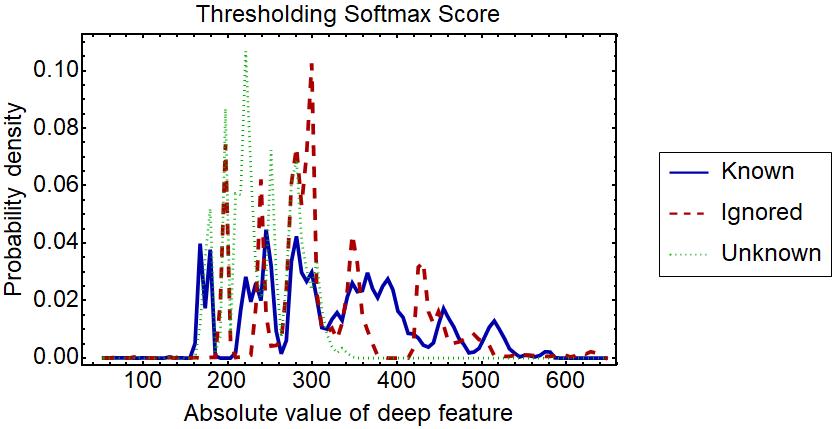}
\end{adjustbox}

\section*{Entropic Open Set and Objectosphere methods}
\label{Obj}
The Entropic Open Set loss function~\cite{dhamija2018reducing} is defined as: 

$$
V_E\left(x\right) = \begin{cases}
- \log\left(S_c\left(x\right)\right), if \ x \in\mathcal{K} \\
-\frac{1}{C}\sum_{c = 1}^C \log\left(S_c\left(x\right)\right), if \ x \in\mathcal{I}
\end{cases}
$$

where $C = 20$ is the number of classes of interest (amino acids). Note, for the $\mathcal{K}$ classes, it reduces to a regular categorical cross entropy loss function, and the $\mathcal{N}$ classes are not involved, since the NN is not aware of them until the testing stage. For the samples belonging to $\mathcal{I}$ classes, $V_E\left(x\right)$ aims to maximize the entropy, and uniformly spread the NN's output over the known classes $\mathcal{K}$. 

The Objectosphere loss function involves the addition of the deep feature $F$, and minimizes its absolute value for the ignored classes and maximizes it for the known ones:

$$
V_O\left(x\right) = V_E\left(x\right) + \alpha
\begin{cases}
max\left(\beta - \lvert\lvert F \rvert\rvert^2, 0 \right), if \ x \in\mathcal{K} \\
\lvert\lvert F \rvert\rvert^2, if \ x \in\mathcal{I}
\end{cases}$$
effectively increasing the separation between the $\mathcal{K}$ and $\{\mathcal{I, N}\}$ classes. The values of $\alpha$ and $\beta$ are adjusted numerically by minimizing the number of inconclusive outcomes on the $\mathcal{K}$ classes. The result of this that $FP = 0\%$ on the $\mathcal{I}$ classes.

Both these approaches provide a significant improvement over the naive approaches as shown in Fig.~\ref{Objresults}. For the entropic open set loss function shown in Fig.~\ref{Objresults}(a), at the cutoff value $\Lambda_{Entropic} = 0.93$, $FP = 0\%$, the NN has $0\%$ wrong outcomes on the known classes, and number of inconclusive outcomes on the known classes is $2.37\%$. For the Objectosphere loss function shown in Fig.~\ref{Objresults}(b), at the cutoff value $\Lambda_{Objectosphere} = 0.91$, $FP = 0\%$, $0\%$ wrong outcomes on the known classes, and the number of inconclusive outcomes on the known classes is drastically reduced to $0.26\%$ as shown in Fig.~\ref{Objresults}. This shows that the Objectosphere reduces inconclusive outcomes by $89\%$ from the entropic open set loss and $96\%$ from the naive approaches, which is a significant improvement and demonstrates that the Objectosphere approach successfully avoids misclassifications and false positives.

The corresponding UMAPs shown in Fig.~\ref{EOSresults} demonstrate a significant improvement in treating the $\mathcal{N}$ classes by the NN. The area taken by the $\mathcal{N}$ classes is significantly reduced in comparison with the previous approaches, one can make a clear distinction between $\mathcal{N}$ and $\mathcal{K}$ classes and draw a line separating them.

\begin{adjustbox}{center,caption={Results of the (a) Entropic Open Set and (b) Objectosphere loss functions. The Objectosphere improves the treatment of the $\mathcal{N}$ classes even more by reducing the number of inconclusive outcomes on the $\mathcal{K}$ classes to $0.26\%$.}, label={Objresults},nofloat=figure,vspace=\bigskipamount}
\includegraphics[width=\textwidth]{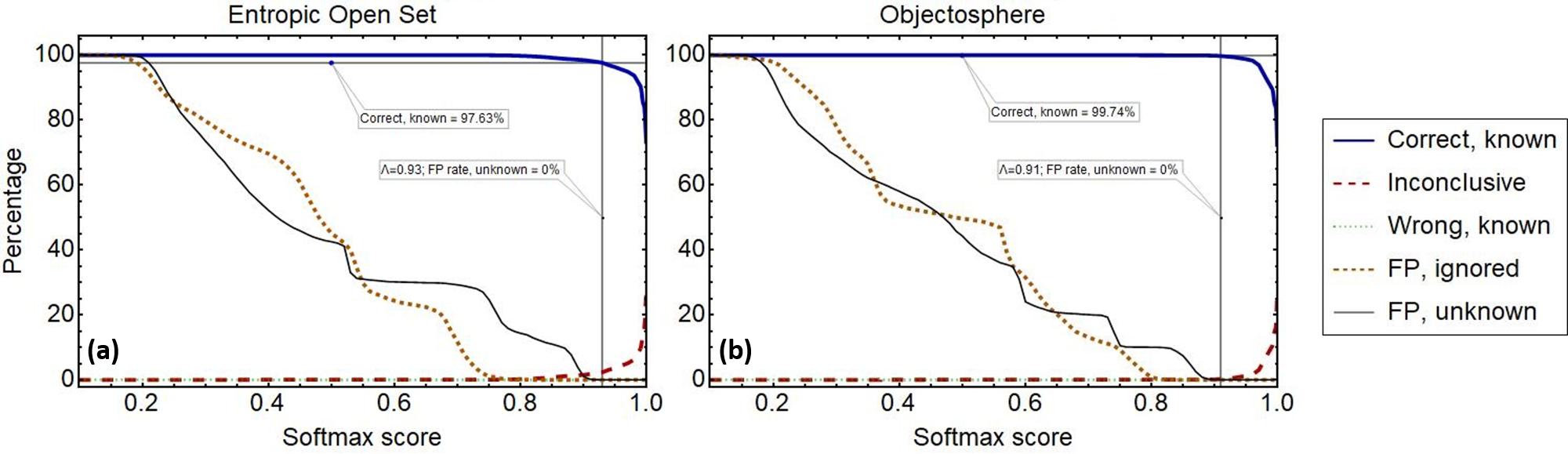}
\end{adjustbox}

\begin{adjustbox}{center,caption={UMAPs for the (a) Entropic Open Set and (b) Objectosphere loss functions. For the Objectosphere case, the area over which the $\mathcal{N}$ classes are scattered shrinks significantly and a clear separation between $\mathcal{K}$ and $\mathcal{N}$ is observed.}, label={EOSresults},nofloat=figure,vspace=\bigskipamount}
\includegraphics[width=\textwidth]{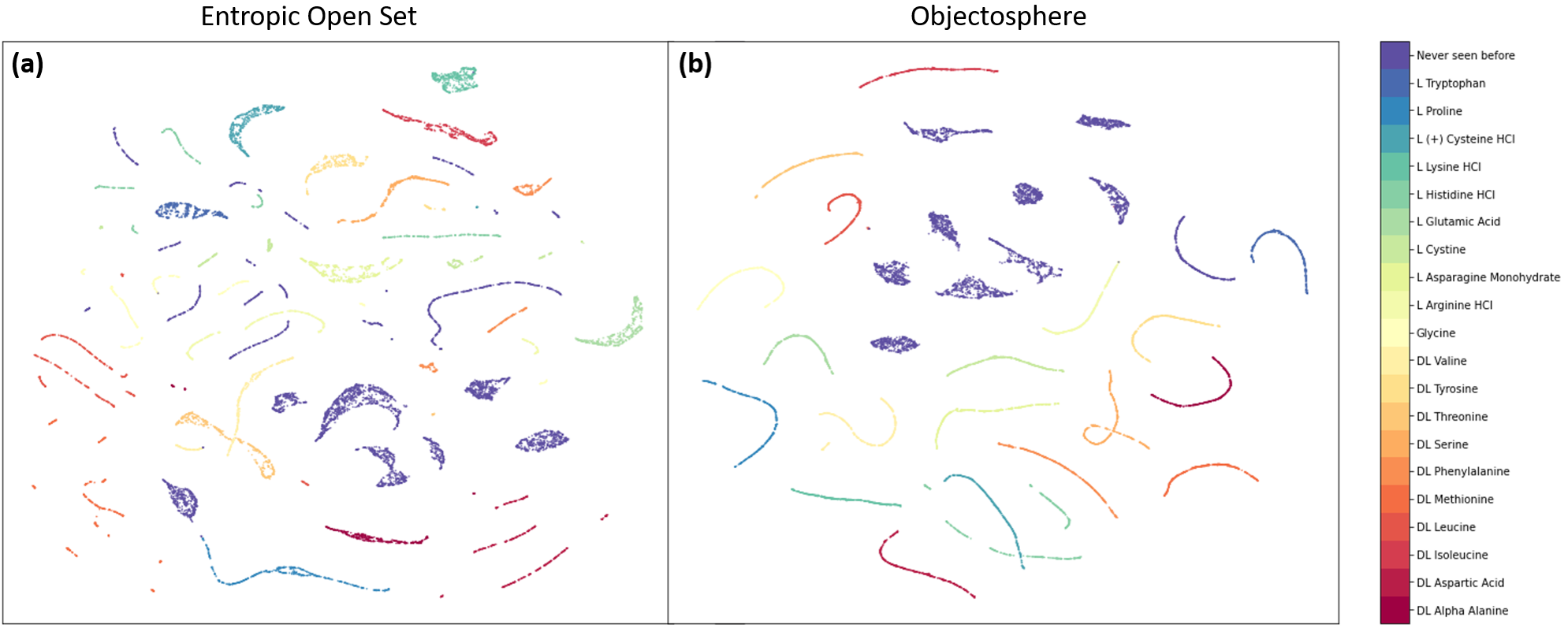}
\end{adjustbox}

Finally, as one can observe in Fig.~\ref{EOSthresh}, the Objectosphere loss function enhances the separation between the features of the $\mathcal{N}$ and $\mathcal{K}$ classes minimizing the incorrect responses from the NN to the new, never seen before inputs.

\begin{adjustbox}{center,caption={Distribution of the deep features of the known ($\mathcal{K}$) and unknown/unexpected inputs ($\mathcal{N}$) by the (a) Entropic Open Set and (b) Objectosphere loss functions. An improved separation of $\mathcal{K}/\mathcal{N}$ classes for the case of Objectosphere is observed.}, label={EOSthresh},nofloat=figure,vspace=\bigskipamount}
\includegraphics[width=\textwidth]{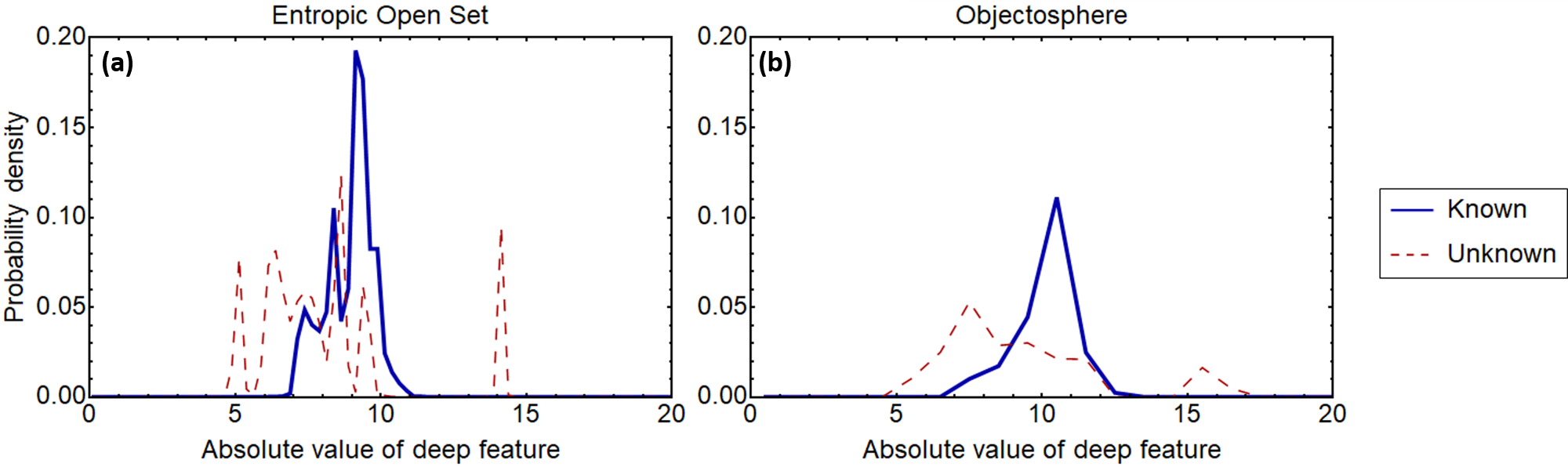}
\end{adjustbox}

\section*{Conclusions and future work}
\label{Conclusions}

Raman spectroscopy in combination with Machine Learning demonstrates significant promise as a fast, label-free, pathogen identification method. It has the potential to reduce treatment costs, mitigate antibiotic resistance, and save human lives. However, traditional ML architectures applied in this context only work in closed-world settings and become ill-defined when presented with unknowns, limiting their practical applications. Here we have shown that we can overcome this challenge by modifying  ResNet26 architecture. The application of the recently introduced entropic open set and Objectosphere approaches enables the ResNet26 architecture to maintain a high accuracy on the known classes while successfully avoiding misclassification of new, never seen before classes. This method is an important step towards the translation of using machine learning methods in clinical applications. In future work we plan to explore mixtures of chemical classes and spectra with low signal-to-noise ratios to better mimic real-world settings and further improve the performance of our neural network. 

\section*{Code availability and reproducibility of our results}
\label{Reproducibility}

\footnote{\href{https://github.com/BalytskyiJaroslaw/RamanOpenSet.git}{https://github.com/BalytskyiJaroslaw/RamanOpenSet.git}}{\href{https://github.com/BalytskyiJaroslaw/RamanOpenSet.git}{}}Our code is publicly available on GitHub. In order to reproduce our results, one needs to upload this data and code into Google Colab~\cite{bisong2019building}, connect to the TPU, and follow the instructions in the Jupyter notebooks provided.

\begin{acknowledgments}

Research reported in this publication was supported by the National Institute of General Medical Sciences of the National Institutes of Health under award No. 1R15GM128166–01. This work was also supported by the UCCS BioFrontiers Center. The funding sources had no involvement in study design; in the collection, analysis, and interpretation of data; in the writing of the report; or in the decision to submit the article for publication. This work was supported in part by the U.S. Civilian Research \& Development Foundation (CRDF Global). The authors would like to thank  Kyle Culhane, Van Hovenga, and Anatoliy Pinchuk for useful discussions.

\end{acknowledgments}

\section*{Author Contributions}
All authors discussed the organization and content of the manuscript. YB developed the modification to the ResNet26 architecture, helped to collect Raman maps of the chemical classes, and wrote the main manuscript. JB and TP helped to collect the Raman maps of the 40 chemical classes. GH supervised the work, assisted in the experimental design and revised the manuscript. KM supervised the work, critically reviewed, and revised the main manuscript. 
\section*{Ethics Declaration}
The authors declare no competing interests.

\pagebreak
\bibliographystyle{unsrt}
\bibliography{apssamp.bib}

\providecommand{\noopsort}[1]{}\providecommand{\singleletter}[1]{#1}%
\begin{thebibliography}{10}

\bibitem{fleischmann2016assessment}
Carolin Fleischmann, Andr{\'e} Scherag, Neill~KJ Adhikari, Christiane~S Hartog,
  Thomas Tsaganos, Peter Schlattmann, Derek~C Angus, and Konrad Reinhart.
\newblock Assessment of global incidence and mortality of hospital-treated
  sepsis. current estimates and limitations.
\newblock {\em American Journal of Respiratory and Critical Care Medicine},
  193(3):259--272, 2016.

\bibitem{deantonio2016epidemiology}
Rodrigo DeAntonio, Juan-Pablo Yarzabal, James~Philip Cruz, Johannes~E Schmidt,
  and Jos Kleijnen.
\newblock Epidemiology of community-acquired pneumonia and implications for
  vaccination of children living in developing and newly industrialized
  countries: A systematic literature review.
\newblock {\em Human Vaccines \& Immunotherapeutics}, 12(9):2422--2440, 2016.

\bibitem{howes2014plasmonic}
Philip~D Howes, Subinoy Rana, and Molly~M Stevens.
\newblock Plasmonic nanomaterials for biodiagnostics.
\newblock {\em Chemical Society Reviews}, 43(11):3835--3853, 2014.

\bibitem{chung2013magneto}
Hyun~Jung Chung, Cesar~M Castro, Hyungsoon Im, Hakho Lee, and Ralph Weissleder.
\newblock A magneto-\uppercase{DNA} nanoparticle system for rapid detection and
  phenotyping of bacteria.
\newblock {\em Nature Nanotechnology}, 8(5):369--375, 2013.

\bibitem{levy2018surviving}
Mitchell~M Levy, Laura~E Evans, and Andrew Rhodes.
\newblock The surviving sepsis campaign bundle: 2018 update.
\newblock {\em Intensive Care Medicine}, 44(6):925--928, 2018.

\bibitem{chaudhuri2008efns}
A~Chaudhuri, PM~Martin, PGE Kennedy, R~Andrew~Seaton, P~Portegies, M~Bojar,
  I~Steiner, and EFNS~Task Force.
\newblock Efns guideline on the management of community-acquired bacterial
  meningitis: report of an efns task force on acute bacterial meningitis in
  older children and adults.
\newblock {\em European Journal of Neurology}, 15(7):649--659, 2008.

\bibitem{brusselaers2011rising}
Nele Brusselaers, Dirk Vogelaers, and Stijn Blot.
\newblock The rising problem of antimicrobial resistance in the intensive care
  unit.
\newblock {\em Annals of Intensive Care}, 1(1):1--7, 2011.

\bibitem{sullivan2001effect}
{\AA}sa Sullivan, Charlotta Edlund, and Carl~Erik Nord.
\newblock Effect of antimicrobial agents on the ecological balance of human
  microflora.
\newblock {\em The Lancet Infectious Diseases}, 1(2):101--114, 2001.

\bibitem{fleming2016prevalence}
Katherine~E Fleming-Dutra, Adam~L Hersh, Daniel~J Shapiro, Monina Bartoces,
  Eva~A Enns, Thomas~M File, Jonathan~A Finkelstein, Jeffrey~S Gerber, David~Y
  Hyun, Jeffrey~A Linder, et~al.
\newblock Prevalence of inappropriate antibiotic prescriptions among us
  ambulatory care visits, 2010-2011.
\newblock {\em JAMA}, 315(17):1864--1873, 2016.

\bibitem{davies2010origins}
Julian Davies and Dorothy Davies.
\newblock Origins and evolution of antibiotic resistance.
\newblock {\em Microbiology and Molecular Biology Reviews}, 74(3):417--433,
  2010.

\bibitem{pang2016review}
Shintaro Pang, Tianxi Yang, and Lili He.
\newblock Review of surface enhanced raman spectroscopic (sers) detection of
  synthetic chemical pesticides.
\newblock {\em TrAC Trends in Analytical Chemistry}, 85:73--82, 2016.

\bibitem{goodfellow2016deep}
Ian Goodfellow, Yoshua Bengio, and Aaron Courville.
\newblock {\em Deep learning}.
\newblock MIT press, 2016.

\bibitem{krizhevsky2012imagenet}
Alex Krizhevsky, Ilya Sutskever, and Geoffrey~E Hinton.
\newblock Imagenet classification with deep convolutional neural networks.
\newblock {\em Advances in Neural Information Processing Systems},
  25:1097--1105, 2012.

\bibitem{farabet2012learning}
Clement Farabet, Camille Couprie, Laurent Najman, and Yann LeCun.
\newblock Learning hierarchical features for scene labeling.
\newblock {\em IEEE Transactions on Pattern Analysis and Machine Intelligence},
  35(8):1915--1929, 2012.

\bibitem{szegedy2015going}
Christian Szegedy, Wei Liu, Yangqing Jia, Pierre Sermanet, Scott Reed, Dragomir
  Anguelov, Dumitru Erhan, Vincent Vanhoucke, and Andrew Rabinovich.
\newblock Going deeper with convolutions.
\newblock In {\em Proceedings of the IEEE conference on Computer Vision and
  Pattern Recognition}, pages 1--9, 2015.

\bibitem{Hagen2021}
Guy~M. Hagen, Justin Bendesky, Rosa Machado, Tram-Ahn Nguyen, Ranman Kumar, and
  Jonathan Ventura.
\newblock Fluorescence microscopy datasets for training deep neural networks.
\newblock {\em Gigascience}, 10(5):giab032, 2021.

\bibitem{dahl2013improving}
George~E Dahl, Tara~N Sainath, and Geoffrey~E Hinton.
\newblock Improving deep neural networks for lvcsr using rectified linear units
  and dropout.
\newblock In {\em 2013 IEEE International Conference on Acoustics, Speech and
  Signal Processing}, pages 8609--8613. IEEE, 2013.

\bibitem{hinton2012deep}
Geoffrey Hinton, Li~Deng, Dong Yu, George~E Dahl, Abdel-Rahman Mohamed, Navdeep
  Jaitly, Andrew Senior, Vincent Vanhoucke, Patrick Nguyen, Tara~N Sainath,
  et~al.
\newblock Deep neural networks for acoustic modeling in speech recognition: The
  shared views of four research groups.
\newblock {\em IEEE Signal Processing Magazine}, 29(6):82--97, 2012.

\bibitem{goodacre1998rapid}
Royston Goodacre, Eadaoin~M Timmins, Rebecca Burton, Naheed Kaderbhai, Andrew~M
  Woodward, Douglas~B Kell, and Paul~J Rooney.
\newblock Rapid identification of urinary tract infection bacteria using
  hyperspectral whole-organism fingerprinting and artificial neural networks.
\newblock {\em Microbiology}, 144(5):1157--1170, 1998.

\bibitem{sigurdsson2004detection}
Sigurdur Sigurdsson, Peter~Alshede Philipsen, Lars~Kai Hansen, Jan Larsen,
  Monika Gniadecka, and Hans-Christian Wulf.
\newblock Detection of skin cancer by classification of raman spectra.
\newblock {\em IEEE Transactions on Biomedical Engineering}, 51(10):1784--1793,
  2004.

\bibitem{olaetxea2020machine}
Ion Olaetxea, Ana Valero, Eneko Lopez, Hector Lafuente, Ander Izeta, Ibon
  Jaunarena, and Andreas Seifert.
\newblock Machine learning-assisted raman spectroscopy for ph and lactate
  sensing in body fluids.
\newblock {\em Analytical Chemistry}, 92(20):13888--13895, 2020.

\bibitem{marigheto1998comparison}
NA~Marigheto, EK~Kemsley, M~Defernez, and RH~Wilson.
\newblock A comparison of mid-infrared and raman spectroscopies for the
  authentication of edible oils.
\newblock {\em Journal of the American Oil Chemists’ Society},
  75(8):987--992, 1998.

\bibitem{ciodaro2012online}
T~Ciodaro, D~Deva, JM~De~Seixas, and D~Damazio.
\newblock Online particle detection with neural networks based on topological
  calorimetry information.
\newblock In {\em Journal of Physics: Conference Series}, volume 368, 1, page
  012030. IOP Publishing, 2012.

\bibitem{liu1993chemometric}
Ying Liu, Belle~R Upadhyaya, and Masoud Naghedolfeizi.
\newblock Chemometric data analysis using artificial neural networks.
\newblock {\em Applied Spectroscopy}, 47(1):12--23, 1993.

\bibitem{gniadecka1997diagnosis}
Monika Gniadecka, HC~Wulf, N~Nymark Mortensen, O~Faurskov Nielsen, and
  DH~Christensen.
\newblock Diagnosis of basal cell carcinoma by raman spectroscopy.
\newblock {\em Journal of Raman Spectroscopy}, 28(2-3):125--129, 1997.

\bibitem{liu2017deep}
Jinchao Liu, Margarita Osadchy, Lorna Ashton, Michael Foster, Christopher~J
  Solomon, and Stuart~J Gibson.
\newblock Deep convolutional neural networks for raman spectrum recognition: a
  unified solution.
\newblock {\em Analyst}, 142(21):4067--4074, 2017.

\bibitem{hochreiter2001gradient}
Sepp Hochreiter, Yoshua Bengio, Paolo Frasconi, J{\"u}rgen Schmidhuber, et~al.
\newblock {\em Gradient flow in recurrent nets: the difficulty of learning
  long-term dependencies}.
\newblock IEEE Press, 2001.

\bibitem{he2016deep}
Kaiming He, Xiangyu Zhang, Shaoqing Ren, and Jian Sun.
\newblock Deep residual learning for image recognition.
\newblock In {\em Proceedings of the IEEE Conference on Computer Vision and
  Pattern Recognition}, pages 770--778, 2016.

\bibitem{ho2019rapid}
Chi-Sing Ho, Neal Jean, Catherine~A Hogan, Lena Blackmon, Stefanie~S Jeffrey,
  Mark Holodniy, Niaz Banaei, Amr~AE Saleh, Stefano Ermon, and Jennifer Dionne.
\newblock Rapid identification of pathogenic bacteria using raman spectroscopy
  and deep learning.
\newblock {\em Nature Communications}, 10(1):1--8, 2019.

\bibitem{maruthamuthu2020raman}
Murali~K Maruthamuthu, Amir~Hossein Raffiee, Denilson~Mendes De~Oliveira,
  Arezoo~M Ardekani, and Mohit~S Verma.
\newblock Raman spectra-based deep learning: A tool to identify microbial
  contamination.
\newblock {\em MicrobiologyOpen}, 9(11):e1122, 2020.

\bibitem{thrift2020deep}
William~John Thrift, Sasha Ronaghi, Muntaha Samad, Hong Wei, Dean~Gia Nguyen,
  Antony~Superio Cabuslay, Chloe~E Groome, Peter~Joseph Santiago, Pierre Baldi,
  Allon~I Hochbaum, et~al.
\newblock Deep learning analysis of vibrational spectra of bacterial lysate for
  rapid antimicrobial susceptibility testing.
\newblock {\em ACS Nano}, 14(11):15336--15348, 2020.

\bibitem{lussier2020deep}
F{\'e}lix Lussier, Vincent Thibault, Benjamin Charron, Gregory~Q Wallace, and
  Jean-Francois Masson.
\newblock Deep learning and artificial intelligence methods for raman and
  surface-enhanced raman scattering.
\newblock {\em TrAC Trends in Analytical Chemistry}, 124:115796, 2020.

\bibitem{peiffer2020machine}
Nathan Peiffer-Smadja, Sarah Delli{\`e}re, Christophe Rodriguez, Gabriel
  Birgand, F-X Lescure, Slim Fourati, and Etienne Rupp{\'e}.
\newblock Machine learning in the clinical microbiology laboratory: has the
  time come for routine practice?
\newblock {\em Clinical Microbiology and Infection}, 26(10):1300--1309, 2020.

\bibitem{lu2020combination}
Weilai Lu, Xiuqiang Chen, Lu~Wang, Hanfei Li, and Yu~Vincent Fu.
\newblock Combination of an artificial intelligence approach and laser tweezers
  raman spectroscopy for microbial identification.
\newblock {\em Analytical Chemistry}, 92(9):6288--6296, 2020.

\bibitem{dhamija2018reducing}
Akshay~Raj Dhamija, Manuel G{\"u}nther, and Terrance~E Boult.
\newblock Reducing network agnostophobia.
\newblock {\em arXiv Preprint arXiv:1811.04110}, 2018.

\bibitem{matan1990handwritten}
Ofer Matan, RK~Kiang, CE~Stenard, B~Boser, JS~Denker, Don Henderson, RE~Howard,
  W~Hubbard, LD~Jackel, and Yann Le~Cun.
\newblock Handwritten character recognition using neural network architectures.
\newblock In {\em 4th USPS Advanced Technology Conference}, volume 2, 5, pages
  1003--1011, 1990.

\bibitem{de2000reject}
Claudio De~Stefano, Carlo Sansone, and Mario Vento.
\newblock To reject or not to reject: that is the question-an answer in case of
  neural classifiers.
\newblock {\em IEEE Transactions on Systems, Man, and Cybernetics, Part C
  (Applications and Reviews)}, 30(1):84--94, 2000.

\bibitem{fumera2002support}
Giorgio Fumera and Fabio Roli.
\newblock Support vector machines with embedded reject option.
\newblock In {\em International Workshop on Support Vector Machines}, pages
  68--82. Springer, 2002.

\bibitem{mcinnes2018umap}
Leland McInnes, John Healy, and James Melville.
\newblock Umap: Uniform manifold approximation and projection for dimension
  reduction.
\newblock {\em arXiv Preprint arXiv:1802.03426}, 2018.

\bibitem{bisong2019building}
Ekaba Bisong.
\newblock {\em Building machine learning and deep learning models on Google
  Cloud Platform}.
\newblock Springer, 2019.

\end{thebibliography}

\end{document}